\documentclass[journal]{IEEEtran}

\ifCLASSINFOpdf

\else

\fi

\usepackage{framed}
\usepackage[ruled]{algorithm2e}
\usepackage{amssymb}
\usepackage{latexsym}
\usepackage{epsfig}
\usepackage{float}
\usepackage{booktabs,multirow, multicol}
\usepackage{cite}
\usepackage{caption}
\usepackage{rotating}
\usepackage{epstopdf}
\usepackage[margin=1in]{geometry}
\usepackage{graphicx}
\usepackage[pagebackref=false,breaklinks=true,colorlinks,bookmarks=false]{hyperref}
\usepackage{color}
\usepackage{subcaption}
\usepackage{amsmath}
\usepackage{verbatim}
\usepackage{gensymb}
\usepackage{array, longtable}
\usepackage{wrapfig}
\usepackage{enumitem}
\usepackage{tabularx}
\usepackage{url}
\usepackage{xcolor}

\usepackage{epsfig}
\usepackage{graphicx}
\usepackage{amsmath}
\usepackage{amssymb}
\usepackage{gensymb}
\usepackage{footmisc}

\usepackage{cite}
\usepackage{array}
\usepackage{fixltx2e}
\usepackage{url}

\usepackage{amssymb}
\usepackage{pifont}
\newcommand{\cmark}{\ding{51}}%
\newcommand{\xmark}{\ding{55}}%

\hypersetup{colorlinks,breaklinks,
            urlcolor=[rgb]{0,0.0,0.0},
            linkcolor=[rgb]{0,0.0,0.0},
            citecolor=[rgb]{0,0.0,0.0}}

\graphicspath{{pics/}}

\begin{document}

\title{Biometric Face Presentation Attack Detection with Multi-Channel Convolutional Neural Network}

\author{Anjith~George, Zohreh~Mostaani, David~Geissenbuhler, Olegs~Nikisins, Andr{\'e}~Anjos
        and~S{\'e}bastien~Marcel
\thanks{A. George, Z. Mostaani, D. Geissenbuhler, O. Nikisins, A. Anjos and S. Marcel are in Idiap Research Institute, Centre du Parc, Rue Marconi 19, CH - 1920, Martigny, Switzerland. }
\thanks{\textcopyright2019 IEEE.  Personal use of this material is permitted.  Permission from IEEE must be obtained for all other uses, in any current or future media, including reprinting/republishing this material for advertising or promotional purposes, creating new collective works, for resale or redistribution to servers or lists, or reuse of any copyrighted component of this work in other works.}
\thanks{DOI: 10.1109/TIFS.2019.2916652}
\thanks{Contact e-mail: anjith.george@idiap.ch}
}

\markboth{}%
{}

\maketitle

\begin{abstract}
Face recognition is a mainstream biometric authentication method. However, vulnerability to presentation attacks (a.k.a spoofing) limits its usability in unsupervised applications. Even though there are many methods available for tackling presentation attacks (PA), most of them fail to detect sophisticated attacks such as silicone masks.
As the quality of presentation attack instruments improves over time, achieving reliable PA detection with visual spectra alone remains very challenging. We argue that analysis in multiple channels might help to address this issue.  In this context, we propose a multi-channel Convolutional Neural Network based approach for presentation attack detection (PAD).
We also introduce the new Wide Multi-Channel presentation Attack (WMCA) database for face PAD which contains a wide variety of 2D and 3D presentation attacks for both impersonation and obfuscation attacks. Data from different channels such as color, depth, near-infrared and thermal are available to advance the research in face PAD. The proposed method was compared with feature-based approaches and found to outperform the baselines achieving an ACER of 0.3\% on the introduced dataset. The database and the software to reproduce the results are made available publicly.

\end{abstract}


\IEEEpeerreviewmaketitle

\section{Introduction}

\IEEEPARstart{B}{iometrics} offers a secure and convenient way for access control. Face biometrics is one of the most convenient
modalities for biometric authentication due to its non-intrusive nature. Even though face recognition systems are reaching human performance in identifying persons in many challenging datasets \cite{learned2016labeled}, most face recognition systems are still vulnerable to presentation attacks (PA), also known as spoofing \footnote{The term spoofing should be deprecated in favour of presentation attacks to comply with the ISO standards.} attacks \cite{marcel2014handbook}, \cite{ISO1}. Merely presenting a printed photo to an unprotected face recognition system could be enough to fool it \cite{anjos2011counter}. Vulnerability to presentation attacks limits the reliable deployment of such systems for applications in unsupervised conditions.

As per the ISO standard \cite{ISO1}, presentation attack is defined as ``a presentation to the biometric data capture subsystem with the goal of interfering with the operation of the biometric system''. Presentation attacks include both `impersonation' as well as `obfuscation' of identity. The `impersonation' refers to attacks in which the attacker wants to be recognized as a different person, whereas in `obfuscation' attacks, the objective is to hide the identity of the attacker. The biometric characteristic or object used in a presentation attack is known as presentation attack instrument (PAI).

\begin{figure}[t]
\centering
        \includegraphics[width=0.95\linewidth]{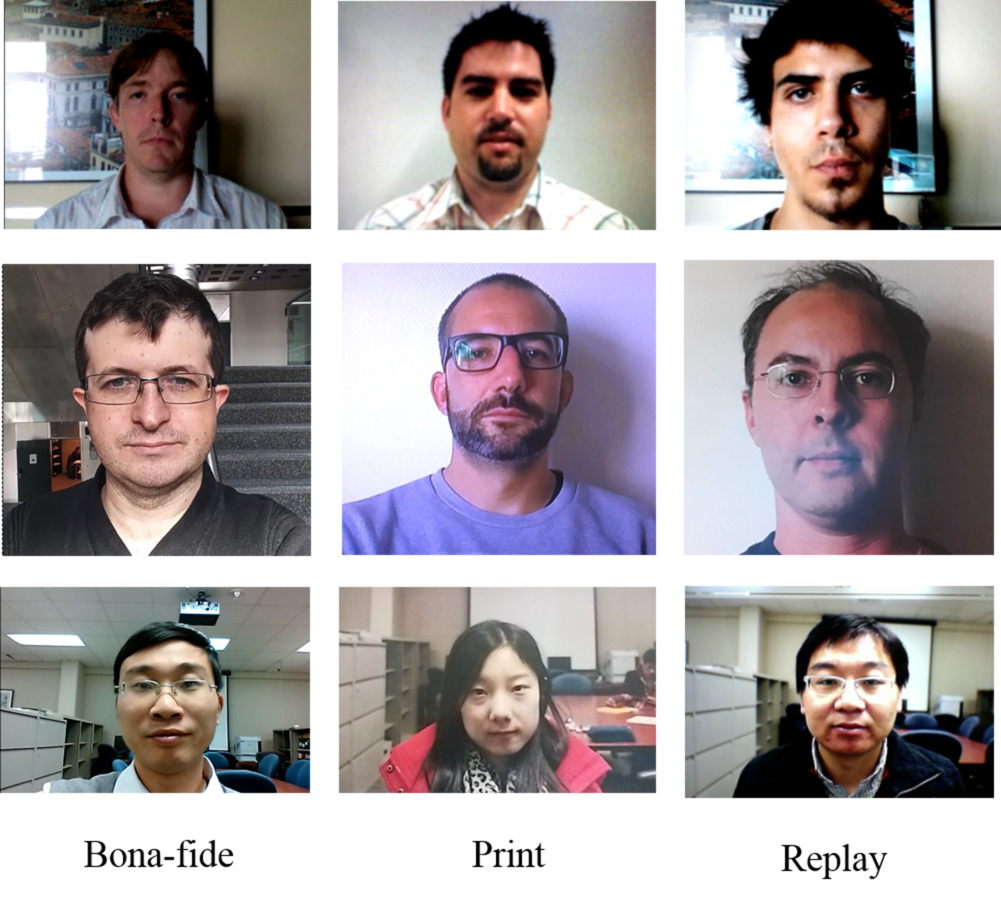}
        \caption{Figure showing \textit{bonafide}, print and replay attacks from different PAD databases, Replay-Attack \cite{chingovska2012effectiveness} (first row), Replay-Mobile \cite{costa2016replay} (second row), and MSU-MFSD \cite{wen2015face} (third row).}\label{fig:example_attacks}

\end{figure}

Different kinds of PAIs can be used to attack face recognition systems.
The presentation of a printed photo, or replaying a video of a subject, are common examples of 2D PAI which have been extensively explored in the available literature. Examples of \textit{bonafide} and 2D PAs
from publicly available databases are shown in Fig. \ref{fig:example_attacks}. More sophisticated attacks could involve manufacturing custom 3D masks which correspond to a target identity for impersonation or to evade identification. For reliable usage of face recognition technology, it is necessary to develop presentation attack detection (PAD) systems to detect such PAs automatically.

The majority of available research deals with the detection of print and replay attacks using visible spectral data. Most of the methods relies on the limitations of PAIs and quality degradation of the recaptured sample. Features such as color, texture \cite{boulkenafet2015face}, \cite{maatta2011face}, motion \cite{anjos2011counter}, and physiological cues \cite{ramachandra2017presentation}, \cite{heusch2019remote} are often leveraged for PAD in images from visible spectrum.

While it is useful to have visible spectrum image based PAD algorithms for legacy face recognition systems, we argue that using only visual spectral information may not be enough for the detection of sophisticated attacks and generalization to new unseen PAIs. The quality of presentation attack instruments (PAI) evolves together with advances in cameras, display devices, and manufacturing methods. Tricking a multi-channel system is harder than a visual spectral one. An attacker would have to mimic real facial features across different representations. The PAD approaches which work in existing PAD databases may not work in real-world conditions when encountered with realistic attacks. Complementary information from multiple channels could improve the accuracy of PAD systems.

The objective of this work is to develop a PAD framework which can detect a variety of 2D and 3D attacks in obfuscation or impersonation settings. To this end, we propose the new Multi-Channel Convolutional Neural Network (MC-CNN) architecture, efficiently combining multi-channel information for robust detection of presentation attacks. The proposed network uses a pre-trained LightCNN model as the base network, which obviates the requirement to train the framework from scratch. In the proposed MC-CNN only low-level LightCNN features across multiple channels are re-trained, while high-level layers of pre-trained LightCNN remain unchanged.

Databases containing a wide variety of challenging PAIs are essential for developing and benchmarking PAD algorithms. In this context, we introduce a Wide Multi-Channel presentation Attack (WMCA) dataset, which contains a broad variety of 2D and 3D attacks. The data split and evaluation protocols are predefined and publicly available. The algorithms, baselines, and the results are reproducible. The software and models to reproduce the results are available publicly \footnote{Source code available at: \url{https://gitlab.idiap.ch/bob/bob.paper.mccnn.tifs2018}}.

The main contributions from this paper are listed below.

\begin{itemize}

\item We propose a novel framework for face presentation attack detection based on multi-channel CNN (MC-CNN). MC-CNN uses a face recognition sub-network, namely LightCNN, making the framework reusable for both PAD and face recognition. The source codes for the network and instructions to train the model are made publicly available allowing to reproduce the findings. We benchmark the proposed method against selected \textbf{reproducible} baseline available in recent publications on the topic~\cite{nikisins2018effectiveness}, as well as reimplementations of recent literature in multi-channel PAD \cite{lucena2017transfer}. We demonstrate that the multi-channel approach is beneficial in both proposed and baseline systems.
\item The new WMCA database is introduced: the subjects in the database are captured using multiple capturing devices/channels, and the MC data is spatially and temporally aligned. The channels present are color, depth, thermal and infrared. The database contains a wide variety of 2D and 3D presentation attacks, specifically, 2D print and replay attacks, mannequins, paper masks, silicone masks, rigid masks, transparent masks, and non-medical eyeglasses.
\end{itemize}

The rest of the paper is organized as follows. Section 2 revisits available literature related to face presentation attack detection. Section 3 presents the proposed approach. The details about the sensors and the dataset are described in section 4. Experimental procedure followed, and the baseline systems are described in Section 5. Extensive testing and evaluations of the proposed approach, along with comparisons with the baselines, discussions, and limitations of the proposed approach are presented in Section 6. Conclusions and future directions are described in Section 7.

\section{Related work} 
Most of the work related to face presentation attack detection addresses detection of 2D attacks, specifically print and 2D replay attacks. A brief review of recent PAD methods is given in this section.

\begin{table*}[t]
\caption{Recent multi-channel face PAD datasets}
\label{tab:multichannel_datasets}
\begin{tabular}{l|p{1.cm}|p{2cm}|p{3cm}|p{3cm}|p{2cm}}
\toprule
Database        & Year                                               & Samples                                                                  & Attacks                                                                                            & Channels          & \multicolumn{1}{c}{\begin{tabular}[c]{@{}c@{}}Synchronous\\  Capture\end{tabular}}                                                        \\ \midrule

3DMAD \cite{3dmad}  & 2013 & 17 subjects                                                              & 3D: Mask attacks                                                                              & Color and depth &\cmark  \\ \hline

I\textsuperscript{2}BVSD \cite{dhamecha2014recognizing}    & 2013 & 75 subjects                                                              & 3D: Facial disguises                                                                             & Color and thermal &---\\ \hline

GUC-LiFFAD \cite{raghavendra2015presentation} & 2015 & 80 subjects                                                              & 2D: Print and replay                                                                               & Light-field imagery &\cmark \\ \hline
MS-Spoof \cite{chingovska2016face}           &2016  & 21 subjects                                               & 2D: Print                                                                                          & Color and NIR (800nm)                                                                &\cmark \\ \hline
BRSU \cite{steiner2016design} &2016 & 50+ subjects                           & 3D: Masks, facial disguise                                                                  & Color \& 4 SWIR bands &\cmark  \\ \hline
EMSPAD \cite{raghavendra2017vulnerability}  &2017  & 50 subjects                                                            & 2D: Print(laser \&Inkjet)                                                                      & 7-band multi-spectral data                                                  & \cmark \\  \hline
MLFP \cite{agarwal2017face}              &2017     & 10 subjects                           & 3D: Obfuscation with latex masks                                                                  & Visible, NIR and thermal bands                                          & \xmark \\ \bottomrule

\end{tabular}
\end{table*}

\subsection{Feature based approaches for face PAD}

For PAD using visible spectrum images, several methods such as detecting motion patterns \cite{anjos2011counter}, color texture, and histogram based methods in different color spaces, and variants of Local Binary Patterns (LBP) in grayscale \cite{boulkenafet2015face} and color images \cite{chingovska2012effectiveness}, \cite{maatta2011face},  have shown good performance. Image quality based features \cite{galbally2014image} is one of the successful feature based methods available in prevailing literature. Methods identifying moir{\'e} patterns \cite{patel2015live}, and image distortion analysis \cite{wen2015face}, use the alteration of the images due to the replay artifacts. Most of these methods treat PAD as a binary classification problem which may not generalize well for unseen attacks~\cite{nikisins2018effectiveness}.

Chingovska \textit{et al}. \cite{chingovska2015use} studied the amount of client-specific information present in features used for PAD. They used this information to build client-specific PAD methods. Their method showed a 50\% relative improvement and better performance in unseen attack scenarios.

Arashloo \textit{et al}. \cite{arashloo2017anomaly} proposed a new evaluation scheme for unseen attacks. Authors have tested several combinations of binary classifiers and one class classifiers. The performance of one class classifiers was better than binary classifiers in the unseen attack scenario. BSIF-TOP was found successful in both one class and two class scenarios. However, in cross-dataset evaluations, image quality features were more useful. Nikisins \textit{et al}. \cite{nikisins2018effectiveness} proposed a similar one class classification framework using one class Gaussian Mixture Models (GMM). In the feature extraction stage, they used a combination of Image Quality Measures (IQM). The experimental part involved an aggregated database consisting of replay attack \cite{chingovska2012effectiveness}, replay mobile \cite{costa2016replay}, and MSU-MFSD \cite{wen2015face} datasets.

Heusch and Marcel \cite{Heuch2018}, \cite{heusch2019remote} recently proposed a method for using features derived from remote photoplethysmography (rPPG).
They used the long term spectral statistics (LTSS) of pulse signals obtained from available methods for rPPG extraction. The LTSS features were combined with SVM for PA detection.
Their approach obtained better performance than state of the art methods using rPPG in four publicly available databases.

\subsection{CNN based approaches for face PAD}

Recently, several authors have reported good performance in PAD using convolutional neural networks (CNN). Gan \textit{et al}. \cite{gan20173d} proposed a 3D CNN based approach, which utilized the spatial and temporal features of the video. The proposed approach achieved good results in the case of 2D attacks, prints, and videos. Yang \textit{et al}. \cite{yang2014learn} proposed a deep CNN architecture for PAD. A preprocessing stage including face detection and face landmark detection is used before feeding the images to the CNN. Once the CNN is trained, the feature representation obtained from CNN is used to train an SVM classifier and used for final PAD task. Boulkenafet \textit{et al}. \cite{boulkenafet2017competition} summarized the performance of the competition on mobile face PAD. The objective was to evaluate the performance of the algorithms under real-world conditions such as unseen sensors, different illumination, and presentation attack instruments. In most of the cases, texture features extracted from color channels performed the best. Li \textit{et al}. \cite{li2018learning} proposed a 3D CNN architecture, which utilizes both spatial and temporal nature of videos. The network was first trained after data augmentation with a cross-entropy loss, and then with a specially designed generalization loss, which acts as a regularization factor. The Maximum Mean Discrepancy (MMD) distance among different domains is minimized to improve the generalization property.

There are several works involving various auxiliary information in the CNN training process, mostly focusing on the detection of 2D attacks. Authors use either 2D or 3D CNN. The main problem of CNN based approaches mentioned above is the lack of training data, which is usually required to train a network from scratch. One commonly used solution is fine-tuning, rather than a complete training, of the networks trained for face-recognition, or image classification tasks. Another issue is the poor generalization in cross-database and unseen attacks tests. To circumvent these issues, some researchers have proposed methods to train CNN using auxiliary tasks, which is shown to improve generalization properties. These approaches are discussed below.

Liu \textit{et al}. \cite{liu2018learning} presented a novel method for PAD with auxiliary supervision. Instead of training a network end-to-end directly for PAD task, they used CNN-RNN model to estimate the depth with pixel-wise supervision and estimate remote photoplethysmography (rPPG) with sequence-wise supervision. The estimated rPPG and depth were used for PAD task. The addition of the auxiliary task improved the generalization capability.

Atoum \textit{et al}. \cite{atoum2017face} proposed a two-stream CNN for 2D presentation attack detection by combining a patch-based model and holistic depth maps.
For the patch-based model, an end-to-end CNN was trained. In the depth estimation, a fully convolutional network was trained using the entire face image. The generated depth map was converted to a feature vector by finding the mean values in the $N \times N$ grid. The final PAD score was obtained by fusing the scores from the patch and depth CNNs.

Shao \textit{et al}. \cite{shao2017deep} proposed a deep convolutional network-based architecture for 3D mask PAD. They tried to capture the subtle differences in facial dynamics using the CNN. Feature maps obtained from the convolutional layer of a pre-trained VGG \cite{Simonyan15} network was used to extract features in each channel. Optical flow was estimated using the motion constraint equation in each channel. Further, the dynamic texture was learned using the data from different channels. The proposed approach achieved an AUC (Area Under Curve) score of 99.99\% in the 3DMAD dataset.

Lucena \textit{et al}. \cite{lucena2017transfer} presented an approach for face PAD using transfer learning from pre-trained models (FASNet). The VGG16 \cite{Simonyan15} architecture which was pre-trained on ImageNet \cite{ILSVRC15} dataset was used as the base network as an extractor, and they modified the final fully connected layers. The newly added fully connected layers in the network were fine-tuned for PAD task. They obtained HTERs of 0\% and 1.20\% in 3DMAD and Replay-Attack dataset respectively.

\subsection{Multi-channel based approaches and datasets for face PAD}

In general, most of the visible spectrum based PAD methods try to detect the subtle differences in image quality when it is recaptured. However, this method could fail as the quality of capturing devices and printers improves. For 3D attacks, the problem is even more severe. As the technology to make detailed masks is available, it becomes very hard to distinguish between \textit{bonafide} and presentation attacks by just using visible spectrum imaging. Several researchers have suggested using multi-spectral and extended range imaging to solve this issue \cite{raghavendra2017extended}, \cite{steiner2016reliable}.

Akhtar \textit{et al}. \cite{akhtar2015biometric} outlines the major challenges and open issues in biometrics concerning presentation attacks. Specifically, in case of face PAD, they discuss a wide variety of possible attacks and possible solutions. They pointed out that sensor-based solutions which are robust against spoofing attempts and which works even in `in the wild' conditions require specific attention. 

Hadid \textit{et al}. \cite{hadid2015biometrics} presented the results from a large scale study on the effect of spoofing on different biometrics traits. They have shown that most of the biometrics systems are vulnerable to spoofing. One class classifiers were suggested as a possible way to deal with unseen attacks. Interestingly, countermeasures combining both hardware (new sensors) and software were recommended as a robust PAD method which could work against a wide variety of attacks.

Raghavendra \textit{et al}. \cite{raghavendra2017extended} presented an approach using multiple spectral bands for face PAD. The main idea is to use complementary information from different bands. To combine multiple bands, they observed a wavelet-based feature level fusion and a score fusion methodology. They experimented with detecting print attacks prepared using different kinds of printers. They obtained better performance with score level fusion as compared to the feature fusion strategy. 

Erdogmus and Marcel \cite{erdogmus2014spoofing} evaluated the performance of a number of face PAD approaches against 3D masks using 3DMAD dataset. This work demonstrated that 3D masks could fool PAD systems easily. They achieved HTER of 0.95\% and 1.27\% using simple LBP features extracted from color and depth images captured with Kinect.

Steiner \textit{et al}. \cite{steiner2016reliable} presented an approach using multi-spectral SWIR imaging for face PAD.  They considered four wavelengths - 935\textit{nm}, 1060\textit{nm}, 1300\textit{nm} and 1550\textit{nm}. In their approach, they trained an SVM for classifying each pixel as a skin pixel or not. They defined a Region Of Interest (ROI) where the skin is likely to be present, and skin classification results in the ROI is used for classifying PAs. The approach obtained 99.28 \% accuracy in per-pixel skin classification.

Dhamecha \textit{et al}. \cite{dhamecha2013disguise} proposed an approach for PAD by combining the visible and thermal image patches for spoofing detection.
They classified each patch as either \textit{bonafide} or attack and used the \textit{bonafide} patches for subsequent face recognition pipeline.

Agarwal \textit{et al}. \cite{agarwal2017face} proposed a framework for the detection of latex mask attacks from multi-channel data, which comprised of visible, thermal and infrared data. The dataset was collected independently for different channels and hence lacks temporal and spatial alignment between the channels.  They have performed experiments using handcrafted features independently in the multiple channels. For PAD, the best performing system was based on redundant discrete wavelet transform (RDWT) and Haralick \cite{haralick1979statistical} features. They computed the features from RDWT decompositions of each patch in a $4 \times 4$ grid and concatenated them to obtain the final feature vector. The computed feature vectors were used with SVM for the PAD task.  From the experiments, it was shown that the thermal channel was more informative as compared to other channels obtaining 15.4\% EER in the frame-based evaluation. However, experiments with fusion could not be performed since the channels were recorded independently.

In \cite{Bhattacharjee:256262} Bhattacharjee \textit{et al}. showed that it is possible to spoof commercial face recognition systems with
custom silicone masks. They also proposed to use the mean temperature of the face region for PAD.

Bhattacharjee \textit{et al}. \cite{bhattacharjee2017you} presented a preliminary study of using multi-channel information for PAD. In addition to visible spectrum images, they considered thermal, near infrared, and depth channels.  They showed that detecting 3D masks and 2D attacks are simple in thermal and depth channels respectively. Most of the attacks can be detected with a similar approach with combinations of different channels, where the features and combinations of channels to use are found using a learning-based approach.

Several multi-channel datasets have been introduced in the past few years for face PAD.
Some of the recent ones are shown in Table \ref{tab:multichannel_datasets}. From the
table it can be seen that the variety of PAIs is limited in most of the available datasets.

\subsection{Discussions}

In general, presentation attack detection in a real-world scenario is challenging. Most of the PAD methods available in prevailing literature try to solve the problem for a limited number of presentation attack instruments. Though some success has been achieved in addressing 2D presentation attacks, the performance of the algorithms in realistic 3D masks and other kinds of attacks is poor.

As the quality of attack instruments evolves, it becomes increasingly difficult to discriminate between \textit{bonafide} and PAs in the visible spectrum alone. In addition, more sophisticated attacks, like 3D silicone masks, make PAD in visual spectra challenging. These issues motivate the use of multiple channels, making PAD systems harder to by-pass.

We argue that the accuracy of the PAD methods can get better with a multi-channel acquisition system. Multi-channel acquisition from consumer-grade devices can improve the performance significantly. Hybrid methods, combining both extended hardware and software could help in achieving good PAD performance in real-world scenarios. We extend the idea of a hybrid PAD framework and develop a multi-channel framework for presentation attack detection.

\section{Proposed method}

A Multi-Channel Convolutional Neural Network (MC-CNN) based approach is proposed for PAD. The main idea is to use the joint representation from multiple channels for PAD, using transfer learning from a pre-trained face recognition network. Different stages of the framework are described in the following.

\subsection{Preprocessing}

\label{subsec:preprocess}

Face detection is performed in the color channel using the MTCNN algorithm \cite{zhang2016joint}. Once the face bounding box is obtained, face landmark detection is performed in the detected face bounding box using Supervised Descent Method (SDM) \cite{xiong2013supervised}. Alignment is accomplished by transforming image, such that the eye centers and mouth center are aligned to predefined coordinates. The aligned face images are converted to grayscale, and resized, to the resolution of $128 \times 128$ pixels.

The preprocessing stage for non-RGB channels requires the images from different channels to be aligned both spatially and temporally with the color channel. For these channels, the facial landmarks detected in the color channel are reused, and a similar alignment procedure is performed. A normalization using  Mean Absolute Deviation (MAD) \cite{leys2013detecting} is performed to convert the range of non-RGB facial images to 8-bit format.

\begin{figure*}[t]
\centering
\includegraphics[width=0.99\textwidth]{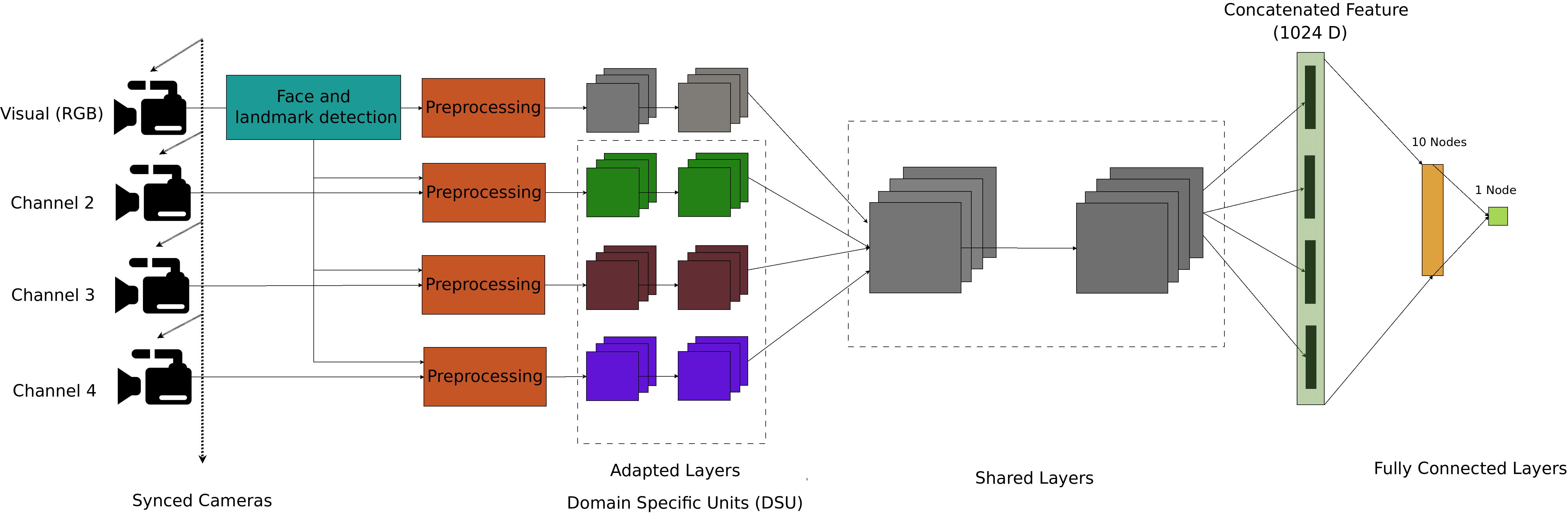}
\caption{Block diagram of the proposed approach. The gray color blocks in the CNN part represent layers which are not retrained, and other colored blocks represent re-trained/adapted layers. }
\label{fig:mcnn_general_block}
\end{figure*}

\subsection{Network architecture}

Many of previous work in face presentation attack detection utilize transfer learning from pre-trained networks. This is required since the data available for PAD task is often of a very limited size, being insufficient to train a deep architecture from scratch. This problem becomes more aggravated when multiple channels of data are involved. We propose a simpler way to leverage a pretrained face recognition model for multi-channel PAD task, adapting a minimal number of parameters. 

The features learned in the low level of CNN networks are usually similar to Gabor filter masks, edges and blobs \cite{yosinski2014transferable}.
Deep CNNs compute more discriminant features as the depth increases \cite{mallat2016understanding}. It has been observed in different studies~\cite{yosinski2014transferable}, \cite{li2015lcnn}, that features, which are closer to the input are more general, while features in the higher levels contain task specific information. Hence, most of the literature in the transfer learning attempts to adapt the higher level features for the new tasks.

Recently, Freitas Pereira \textit{et al}. \cite{freitas2018heterogeneous} showed that the high-level features in deep convolutional
neural networks, trained in visual spectra, are domain independent, and they can be used to encode face images collected from different
image sensing domains. Their idea was to use the shared high-level features for heterogeneous face recognition task, retraining
only the lower layers. In their method they split the parameters of the CNN architecture into two, the higher level features are shared among the different channels, and the lower level features  (known as Domain Specific Units (DSU)) are adapted separately for different modalities. The objective was to learn the same face encoding for different channels, by adapting the DSUs only. The network was trained using contrastive loss (with Siamese architecture) or triplet loss. Re-training of only low-level features has the advantage of modifying a minimal set of parameters.

We extend the idea of domain-specific units (DSU) for multi-channel PAD task. Instead of forcing the representation from different channels to be the same, we leverage the complementary information from a joint representation obtained from multiple channels. We hypothesize that the joint representation contains discriminatory information for PAD task. By concatenating the representation from different channels, and using fully connected layers, a decision boundary for the appearance of \textit{bonafide} and attack presentations can be learned via back-propagation. The lower layer features, as well as the higher level fully connected layers, are adapted in the training phase.

The main idea used from \cite{freitas2018heterogeneous} is the adaptation of lower layers of CNN, instead of adapting the whole network when limited amount of target data is available. The network in \cite{freitas2018heterogeneous} only has one forward path, whereas in MC-CNN the network architecture itself is extended to accommodate multi-channel data. The main advantage of the proposed framework is the adaptation of a minimal amount of network weights when the training data is limited, which is usually the case with available PAD datasets. The proposed framework introduces a new way to deal with multi-channel PAD problem, reusing a large amount of face recognition data available when a limited amount of data is available for training PAD systems. 

In this work, we utilize a LightCNN model \cite{wu2018light}, which was pre-trained on a large number of face images for face recognition. The LightCNN network is especially interesting as the number of parameters is much smaller than in other networks used for face recognition. LightCNN achieves a reduced set of parameters using a Max-Feature Map (MFM) operation as an alternative to Rectified Linear Units (ReLU), which suppresses low activation neurons in each layer.

The block diagram of the proposed framework is shown in Fig. \ref{fig:mcnn_general_block}. The pre-trained LightCNN model produces a 256-dimensional embedding, which can be used as face representation. The LightCNN model is extended to accept four channels. The 256-dimensional representation from all channels are concatenated, and two fully connected layers are added at the end for PAD task. The first fully connected layer has ten nodes, and the second one has one node. Sigmoidal activation functions are used in each fully connected layer. The higher level features are more related to the task to be solved. Hence, the fully connected layers added on top of the concatenated representations are tuned exclusively for PAD task. Reusing the weights from a network pre-trained for face recognition on a large set of data, we avoid plausible over-fitting, which can occur due to a limited amount of training data.

Binary Cross Entropy (BCE) is used as the loss function to train the model using the ground truth information for PAD task.

The equation for BCE is shown below.
\begin{equation}
\mathcal{L}=-{(y\log(p) + (1 - y)\log(1 - p))}
\end{equation}
where $y$ is the ground truth, ($y=0$ for attack and $y=1$ for \textit{bonafide}) and $p$ is predicted probability.

Several experiments were carried out by adapting the different groups of layers, starting from the low-level features. The final fully connected layers are adapted for PAD task in all the experiments.

While doing the adaptation, the weights are always initialized from the weights of the pre-trained layers. Apart from the layers adapted, the parameters for the rest of the network remain shared.

The layers corresponding to the color channel are not adapted since the representation from the color channel can be reused for face recognition, hence making the framework suitable for simultaneous face recognition and presentation attack detection.

\section{The Wide Multi-Channel Presentation Attack Database}

The Wide Multi-Channel presentation Attack (WMCA) database consists of short video recordings for both  \textit{bonafide} and presentation attacks from 72 different identities. In this section, we provide the details on the data collection process, and statistics of the database.

\subsection{Camera set up for data collection}

For acquisition of face data, different sensors were selected to provide a sufficient range of high-quality information in both visual and infrared spectra. In addition, 3D information was provided by one sensor adding a depth map channel to the video stream. Overall, the data stream is composed of a standard RGB video stream, a depth stream (called RGB-D when considered together with the color video), a Near-Infrared (NIR) stream, and a thermal stream. While the RGB-D and NIR data are provided by an Intel RealSense SR300 sensor, the thermal data is provided by a Seek Thermal Compact PRO camera, both being relatively cheap devices aimed at the consumer market. The hardware specifications of these devices are described below.

\subsubsection{Intel RealSense SR300 sensor}

The Intel RealSense SR300 camera is a consumer grade RGB-D sensor aimed at gesture recognition and 3D scanning (Fig.  \ref{fig:idiap_setup}(c)). It features a full-HD RGB camera, capable of capturing resolution of $1920\times1080$ pixels in full-HD mode at 30 frames-per-second (fps) or $1260\times720$ pixels in HD mode at 60 fps. 

\subsubsection{Seek Thermal Compact PRO sensor}

The Seek Thermal Compact PRO sensor is a commercial thermal camera aimed at the consumer market (Fig. \ref{fig:idiap_setup}(b)). It provides a QVGA resolution of $320\times240$. This camera range is primarily intended to be mounted on smart-phone devices. The sensor is capable of capturing at approximately 15 fps, with a non-continuous operation due to an electro-mechanical shutter used to calibrate the sensor at regular intervals (approx. $2s$).

\subsection{Camera integration and calibration}

\begin{figure}[h]
    \begin{center}
        \includegraphics[width=0.45\textwidth]{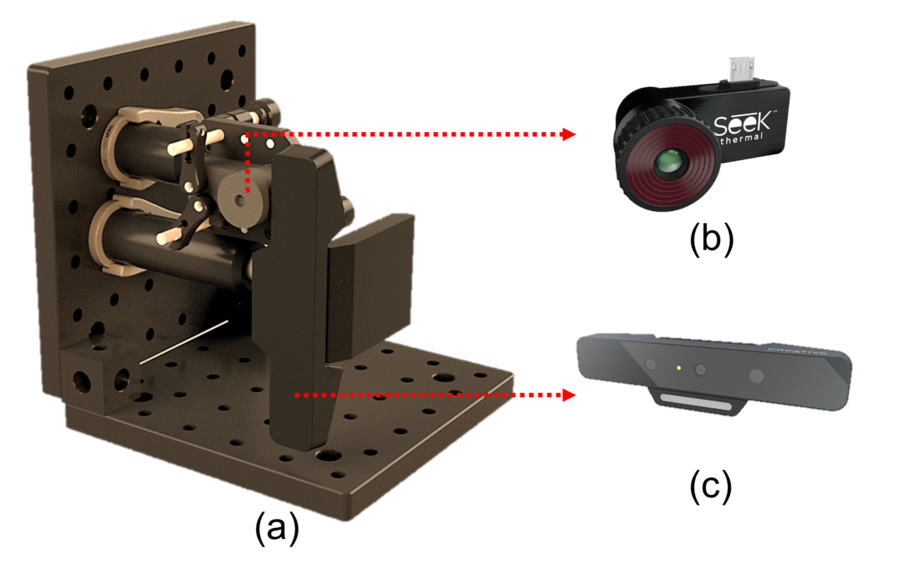}
        \caption{The integrated setup used for WMCA data collection; a) rendering of the integrated system, b) Seek Thermal Compact PRO sensor , c) Intel RealSense SR300 sensor. }\label{fig:idiap_setup}
    \end{center}
\end{figure}

For multi-sensor capture, it is essential that all sensors are firmly attached to a single mounting frame to maintain alignment and minimize vibrations. The setup was built using standard optical mounting posts,  giving an excellent strong and modular mounting frame with the ability to precisely control the orientation of the devices, Fig.~\ref{fig:idiap_setup}.

\begin{figure}[ht]
    \centering
    \includegraphics[width=1\linewidth]{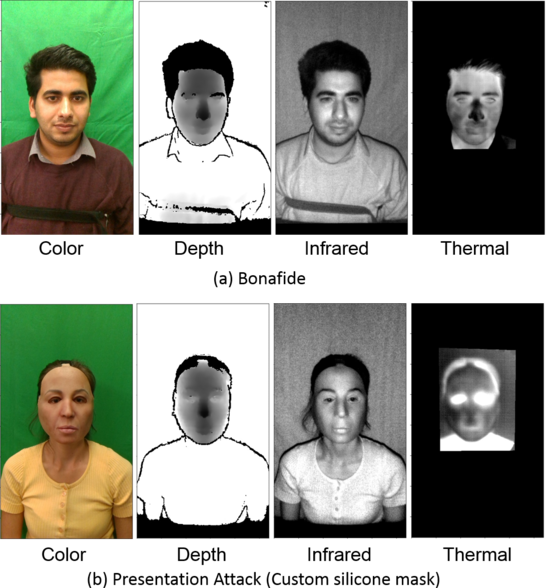}
    \caption{Sample images of a) Bonafide and b) Silicone mask attack from the database for all channels after alignment. The images from all channels are aligned with the calibration parameters and normalized to eight bit
    for better visualization.}
    \label{fig:db_sample}
\end{figure}

To calibrate the cameras and to provide relative alignment of the sensors to the software architecture, we used a checkerboard pattern made from materials with different thermal characteristics. The data from this checkerboard was captured simultaneously from all the channels. For the pattern to be visible on the thermal channel, the target was illuminated by high power halogen lamps. Custom software was then implemented to automatically extract marker points allowing precise alignment of the different video streams. Sample images from all the four channels after alignment is shown in Fig. \ref{fig:db_sample}.

\subsection{Data collection procedure}

\begin{figure}[ht]
    \centering
    \includegraphics[width=.7\linewidth]{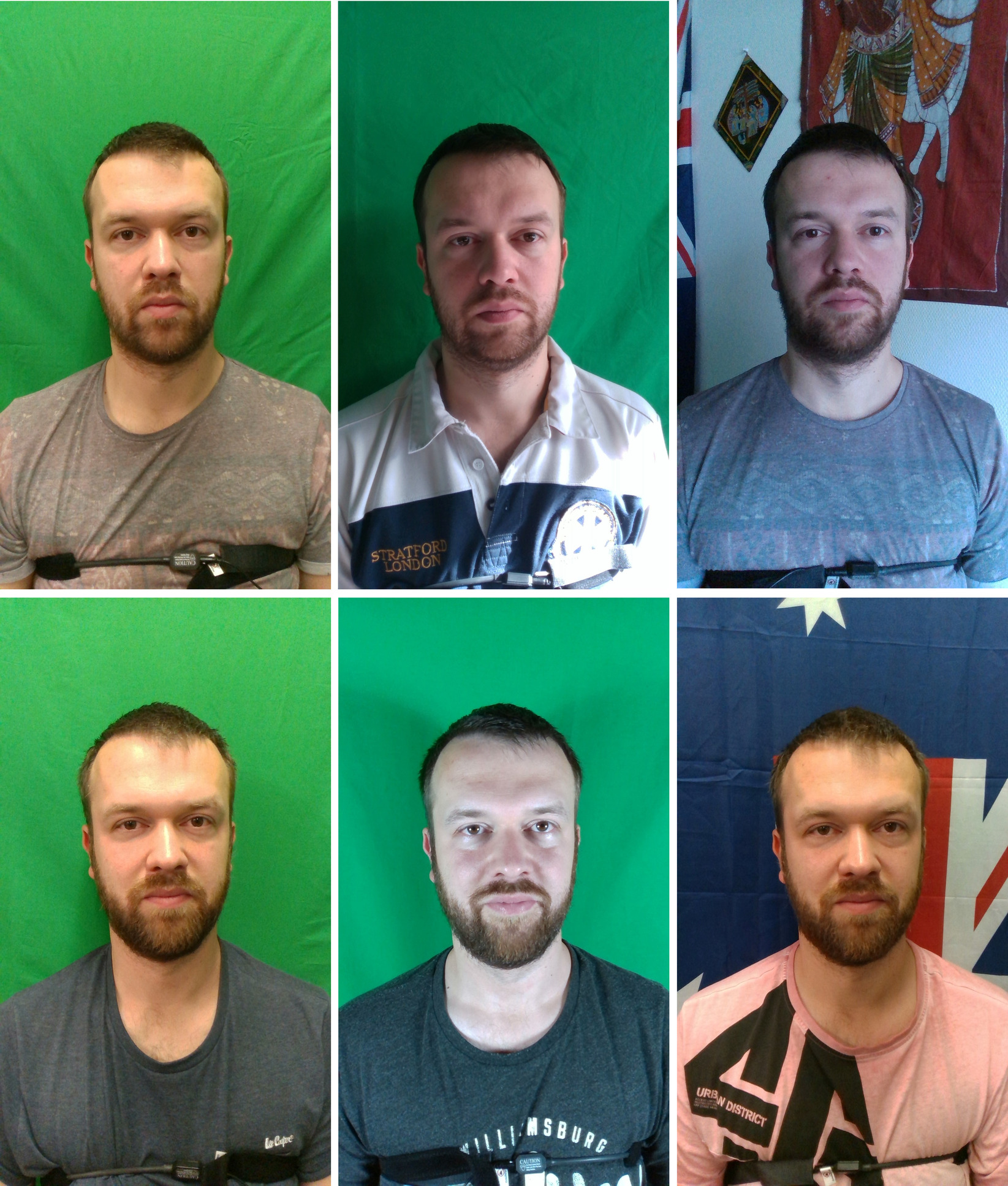}
    \caption{Examples of \textit{bonafide} data in 6 different sessions. Top left is session one and bottom right is session seven. There is no \textit{bonafide} data for session four.}
    \label{fig:bf_rgb}
\end{figure}

The data was acquired during \textbf{seven} sessions over an interval of five months. The sessions were different (Fig. \ref{fig:bf_rgb}) in their environmental conditions such as  background (uniform and complex) and illumination (ceiling office light, side LED lamps, and day-light illumination) (Table \ref{tab:session-info}).
At each session, 10 seconds of data both for \textit{bonafide} and at least two presentation attacks performed by the study participant was captured. Session four was dedicated to presentation attacks only.

\begin{table}[ht]
\centering
\caption{Session description for WMCA data collection}
\label{tab:session-info}
\begin{tabular}{lcr}
\specialrule{.1em}{.05em}{.05em}
\textbf{Session} & \textbf{Background} & \textbf{Illumination} \\
\specialrule{.1em}{.05em}{.05em}
1                & uniform       & ceiling office light                         \\ \hline
2                & uniform       & day-light illumination                         \\ \hline
3                & complex       & day-light illumination                         \\ \hline
4                & uniform       & ceiling office light                         \\ \hline
5                & uniform       & ceiling office light                         \\ \hline
6                & uniform       & side illumination with LED lamps \\ \hline
7                & complex             & ceiling office light                         \\
\specialrule{.1em}{.05em}{.05em}
\end{tabular}
\end{table}

Participants were asked to sit in front of the custom acquisition system and look towards the sensors with a neutral facial expression. If the subjects wore prescription glasses, their \textit{bonafide} data was captured twice, with and without the medical glasses. The masks and mannequins were heated using a blower prior to capture to make the attack more challenging. 
The distance between the subject and the cameras was approximately 40\textit{cm} for both \textit{bonafide} and presentation attacks. 
 The acquisition operator adjusted the cameras so that the subject's face was frontal and located within the field of view of all the sensors at the desired distance. Then they launched the capturing program which recorded data from the sensors for 10 seconds.

\subsection{Presentation attacks}
The presentation attacks were captured under the same conditions as \textit{bonafide}. More than \textbf{eighty} different presentation attack instruments (PAIs) were used for the attacks most of which were presented both by a study participant and on a fixed support.

\begin{figure}[ht]
    \centering
    \includegraphics[width=1\linewidth]{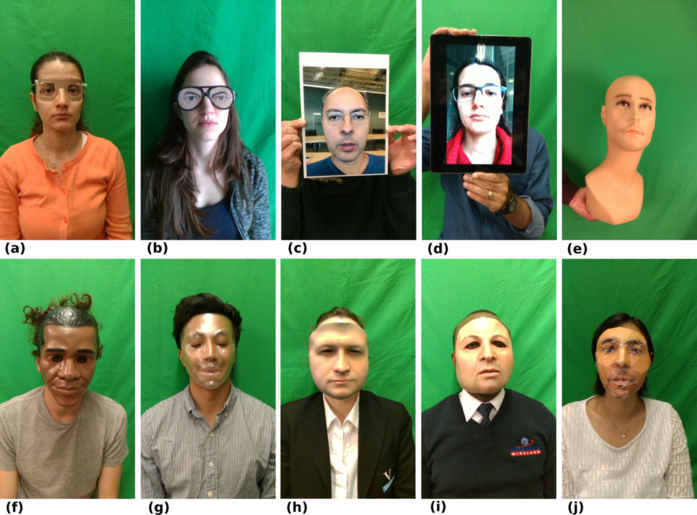}
    \caption{Examples of presentation attacks with different PAIs. (a): glasses (paper glasses), (b): glasses (funny eyes glasses), (c): print, (d): replay, (e): fake head, (f): rigid mask (Obama plastic Halloween mask), (g): rigid mask (transparent plastic mask), (h): rigid mask (custom made realistic), (i): flexible mask (custom made realistic), and (j): paper mask.}
    \label{fig:pais}
\end{figure}

The PAIs presented in this database can be grouped into seven categories. Some examples can be seen in Fig. \ref{fig:pais}.

\begin{itemize}
\item \textbf{glasses:} Different models of disguise glasses with fake eyes (funny eyes glasses) and paper glasses. These attacks constitute partial attacks.

\item \textbf{fake head:} Several models of mannequin heads were used, some of the mannequins were heated with a blower prior to capture.

\item \textbf{print:}  Printed face images on A4 matte and glossy papers using professional quality Ink-Jet printer (Epson\_XP-860) and typical office laser printer (CX c224e). The images were captured by the rear camera of an ``iPhone S6'' and re-sized so that the size of the printed face is human like.

\item \textbf{replay:} Electronic photos and videos. An ``iPad pro 12.9in'' was used for the presentations. The videos were captured in HD at 30 fps by the front camera of an ``iPhone S6'' and in full-HD at 30 fps by the rear camera of the ``iPad pro 12.9in''. Some of the videos were re-sized so that the size of the face presented on the display is human like.

\item \textbf{rigid mask:} Custom made realistic rigid masks and several designs of decorative plastic masks.

\item \textbf{flexible mask:} Custom made realistic soft silicone masks.

\item \textbf{paper mask:} Custom made paper masks based on real identities. The masks were printed on the matte paper using both printers mentioned in the print category.

\end{itemize}

The total number of presentations in the database is 1679,  which contains 347  \textit{bonafide} and 1332 attacks. More detailed information can be found in Table \ref{tab:BATL-data} \footnote{ For downloading the dataset, visit \url{https://www.idiap.ch/dataset/wmca} }.

Each file in the dataset contains data recorded at 30 fps for 10 seconds amounting to 300 frames per channel, except for thermal channel, which contains approximately 150 frames captured at 15 fps. All the channels are recorded in uncompressed format, and the total size of the database is 5.1 TB.

\begin{table}[ht]
\centering
\caption{Statistics for WMCA database.}
\label{tab:BATL-data}
\begin{tabular}{lr}
\specialrule{.1em}{.05em}{.05em}
\textbf{Type}                                   & \#Presentations \\
\specialrule{.1em}{.05em}{.05em}
\textit{bonafide}                                      & 347                      \\ \hline
glasses                                         & 75                       \\ \hline
fake head                                       & 122                      \\ \hline
print                                          & 200                      \\ \hline
replay                                          & 348                      \\ \hline
rigid mask                                      & 137                      \\ \hline
flexible mask                                   & 379                       \\ \hline
paper mask                                      & 71                       \\
\specialrule{.1em}{.05em}{.05em}
\textbf{TOTAL}                                  & \textbf{1679} \\
\specialrule{.1em}{.05em}{.05em}
\end{tabular}
\end{table}

\section{Experiments}

This section describes the experiments performed on the WMCA dataset.  All four channels of data obtained from Intel RealSense SR300 and Seek Thermal Compact PRO are used in the experiments. Various experiments were done to evaluate the performance of the system in ``seen'' and ``unseen''  attack scenarios. In the ``seen'' attack protocol, all types of PAIs are present in the train, development and testing subsets (with disjoint client ids in each fold). This protocol is intended to test the performance of the algorithm in the cases where the attack categories are known a priori. The ``unseen'' attack protocols try to evaluate the performance of the system on PAIs which were not present in the training and development subsets. The ``unseen'' attack protocols thus emulate the realistic scenario of encountering an attack which was not present in the training set. The evaluation protocols are described below.

\subsection{Protocols}

 As the consecutive frames are correlated, we select only 50 frames from each video which are uniformly sampled in the temporal domain. Individual frames from a video are considered as independent samples producing one score per frame. A biometric sample consists of frames from all four channels which are spatially and temporally aligned.

Two different sets of protocols were created for the WMCA dataset.

\begin{itemize}

  \item \textbf{grandtest} protocol : The WMCA dataset is divided into three partitions: $train$, $dev$, and $eval$. The data split is done ensuring almost equal distribution of PA categories and disjoint set of client identifiers in each set. Each of the PAIs had different client id. The split is done in such a way that a specific PA instrument will appear in only one set. The ``grandtest'' protocol emulates the ``seen'' attack scenario as the PA categories are distributed uniformly in the three splits.

  \item \textbf{unseen attack} protocols: The unseen attack protocols defined in the WMCA dataset contains three splits similar to the grandest protocol. Seven unseen attack sub-protocols were created for conducting unseen attack experiments using leave one out (LOO) technique. In each of the unseen attack protocols, one attack is left out in the $train$ and $dev$ sets. The $eval$ set contains the \textit{bonafide} and the samples from the attack which was left out in training. For example, in ``LOO\_fakehead'' protocol, the fake head attacks are not present in both $train$ and $dev$ sets. In the $test$ set, only \textit{bonafide} and fake head attacks were present. The training and tuning are done on data which doesn't contain the attack of interest.
\end{itemize}

\subsection{Evaluation metrics}

We train the framework using the data in the $train$ set, and the decision threshold is found from the $dev$ set by minimizing a given criteria (here, we used a BPCER = 1\% for obtaining the thresholds). We also report the standardized ISO/IEC 30107-3 metrics \cite{ISO}, Attack Presentation Classification Error Rate (APCER),  and Bonafide Presentation Classification Error Rate (BPCER) in the $test$ set at the previously optimized threshold.

To summarize the performance in a single number, the Average Classification Error Rate (ACER) is used, which is an average of APCER and BPCER. The ACER is reported for both $dev$ and $test$ sets.

Apart from the error metrics, ROC curves are also shown for the baseline and MCCNN method.

\subsection{Baseline experiment setup}

Since a new database is proposed in this paper, baseline experiments are performed for each channel first. The selected baselines are reproducible and have either open-source implementation~\cite{nikisins2018effectiveness} or re-implementation. Three sets of baselines are described in this section.

\subsubsection{IQM-LBP-LR baseline}

This baseline consists of Image Quality Measures (IQM)~\cite{galbally2014image} for the RGB channel, and different variants of Local Binary Patterns (LBP) for non-RGB channels. The detailed description of the baseline systems is given below.

An individual PAD algorithm is implemented for every data stream from the camera. The structure of all PAD algorithms can be split into three blocks: preprocessor, a feature extractor, and classifier. The final unit of the PAD system is a fusion block, combining the outputs of channel-specific PAD algorithms, and producing a final decision.

The preprocessing part is exactly similar to the description in Section \ref{subsec:preprocess}, except for color channel. For the color channel, all three RGB channels are retained in the baseline experiments.

The feature extraction step aims to build a discriminative feature representation. For the color channel, the feature vector is composed of 139 IQMs \cite{nikisins2018effectiveness}. Spatially enhanced histograms of LBPs are selected as features for infrared, depth, and thermal channels \cite{nikisins2018effectiveness}. Optimal LBP parameters have been selected experimentally using grid search for each channel independently.

For classification, Logistic Regression (LR) is used as a classifier for color, infrared, depth, and thermal channels \footnote{ We are pointing out that we investigated other classifiers such as SVM but as no performance improvement was noticed we decided to keep a simple method.}. The features are normalized to zero mean and unity standard deviation before the training. The normalization parameters are computed using samples of \textit{bonafide} class only. In the prediction stage, a probability of a sample being a \textit{bonafide} class is computed given trained LR model.

Scores from all PAD algorithms are normalized to $[0, 1]$ range, and a mean fusion is performed to obtain the final PA score.

\subsubsection{RDWT-Haralick-SVM baseline}

In this baseline we used the re-implementation of the algorithm in \cite{agarwal2017face} for individual channels. We applied similar preprocessing strategy as discussed in the previous section in all channels. Haralick \cite{haralick1979statistical} features computed from the RDWT decompositions in a $4 \times 4$ grid are concatenated and fed to a Linear SVM for obtaining the final scores. Apart from implementing the pipeline independently for each channel, we additionally performed a mean fusion of all channels.

\subsubsection{FASNet baseline}

We also compare our system to a deep learning based FASNet \cite{lucena2017transfer} baseline. The FASNet
uses the aligned color images as input for PAD task. We reimplemented the approach in PyTorch \cite{paszke2017automatic} which is made available publicly. 

\subsection{Experiment setup with the proposed MC-CNN approach}

The architecture shown in Fig. \ref{fig:mcnn_general_block} was used for the experiments. The base network used is LightCNN with 29 layers. Further, to accommodate four channels of information, the same network is extended. The embedding layers from different channels are concatenated, and two fully connected layers are added at the end. We performed different experiments by re-training different sets of low-level layers. The layers which are not re-trained are shared across the channels.

The dataset contained an unequal number of samples for \textit{bonafide} and attacks in the training set. The effect of this class imbalance is handled by using a weighted BCE loss function. The weights for the loss function is computed in every mini-batch dynamically based on the number of occurrences of classes in the mini-batch. To compensate for the small amount of training data, we used data augmentation by randomly flipping the images horizontally. All channels are flipped simultaneously to preserve any cross-channel dependencies. A probability of 0.5 was used in this data augmentation. The network was trained using Binary Cross Entropy (BCE) loss using Adam Optimizer \cite{kingma2014adam} with a learning rate of $1\times10^{-4}$. The network was trained for 25 epochs on a GPU grid with a mini-batch size of 32. Implementation was done in PyTorch \cite{paszke2017automatic}.

\section{Results and Discussion}


\subsection{Experiments with \textbf{grandtest} protocol}

\subsubsection{Baseline results}
The performance of the baselines in different individual channels and results from fusion are shown in Table \ref{tab:baseline_results}. From Table \ref{tab:baseline_results}, it can be seen that for individual channels, thermal and infrared provides
more discriminative information. RDWT-Haralick-SVM in infrared channel provides the best accuracy among individual channels. 
It is observed that the fusion of multiple channels improves accuracy. Score fusion improves the accuracy of both feature-based baselines. Deep learning-based baseline FASNet achieves better accuracy as compared to IQM and RDWT-Haralick features in color channel. The FASNet architecture was designed exclusively for three channel color images since it uses normalization parameters and weights from a pre-trained model trained on ImageNet \cite{ILSVRC15} dataset. The training stage in FASNet is performed by fine-tuning the last fully connected layers. The usage of three channel images and finetuning of only last fully connected layers limits the straight forward extension of this architecture to other channels. In the baseline experiments, score fusion of individual channels achieved the best performance and is used as the baselines in the subsequent experiments. From this set of experiments, it is clear that the addition of multiple channels helps in boosting the performance of PAD systems. However, the performance achieved with the best baseline systems is not adequate for deployment in critical scenarios. The lower accuracy in the fusion baselines points to the necessity to have methods which utilize multi-channel information more efficiently.

\begin{table*}[t]
\centering
\caption{Performance of the baseline systems and the components in \textbf{grandtest} protocol of WMCA dataset. The values reported are obtained with a threshold computed for BPCER 1\% in $dev$ set.}
\label{tab:baseline_results}
\begin{tabular}{@{}ccc|ccc@{}}
\toprule
\multirow{2}{*}{Method} & \multicolumn{2}{c|}{dev (\%)}     & \multicolumn{3}{c}{test (\%)}                                  \\ \cmidrule(l){2-6}
                        & \multicolumn{1}{c|}{APCER} & ACER & \multicolumn{1}{c|}{APCER} & \multicolumn{1}{c|}{BPCER} & ACER  \\ \midrule

Color (IQM-LR)         &76.58  &38.79  &87.49     &0  &43.74\\
Depth (LBP-LR)         &57.71  &29.35  &65.45  &0.03  &32.74\\
Infrared (LBP-LR)       &32.79  &16.9  &29.39  &1.18  &15.28\\
Thermal (LBP-LR)        &11.79  &6.4  &16.43   &0.5  &8.47 \\
Score fusion (IQM-LBP-LR Mean fusion)  &10.52  &5.76  &13.92  &1.17  &7.54 \\ \hline
Color (RDWT-Haralick-SVM) &36.02  &18.51  &35.34  &1.67  &18.5\\
Depth (RDWT-Haralick-SVM) &34.71  &17.85  &43.07  &0.57  &21.82\\
Infrared (RDWT-Haralick-SVM)&14.03  &7.51  &12.47  &0.05  &6.26 \\
Thermal (RDWT-Haralick-SVM) &21.51  &11.26  &24.11  &0.85  &12.48 \\
Score fusion (RDWT-Haralick-SVM Mean fusion)  &6.2  &3.6  &6.39  &0.49  &3.44 \\\hline
FASNet &18.89  &9.94  &17.22  &5.65  &11.44 \\ \bottomrule
\end{tabular}
\end{table*}


\subsubsection{Results with MC-CNN}

\begin{figure}[t]
\centering
\includegraphics[width=0.95\linewidth,page=2]{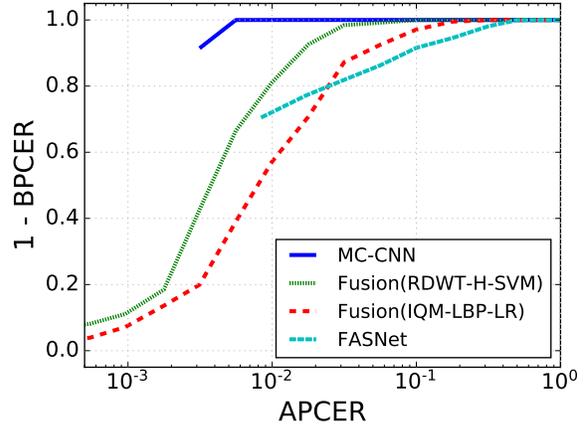}
\caption{ROC for MC-CNN and the baseline methods in WMCA \textbf{grandtest} protocol $eval$ set \textsuperscript{\ref{note1}}.} 
\label{fig:roc_idiap_mccnn}
\end{figure}

The results with the proposed approach in the grandest protocol are shown in Table \ref{tab:perf_mccn}. The corresponding ROCs are shown in Fig. \ref{fig:roc_idiap_mccnn} \footnote{\label{note1} The score distributions from the CNN is bimodal (with low variance in each mode) with most of the values concentrated near zero and one, which explains the lack of points in the lower APCER values in the ROC plots.}.
From Table \ref{tab:perf_mccn}, it can be seen that the proposed MC-CNN approach outperforms the selected baselines by a big margin. From the baseline results, it can be seen that having multiple channels alone doesn't solve the PAD problem. Efficient utilization of information from the various channels is required for achieving good PAD performance. The proposed framework utilizes complementary information from multiple channels with the joint representation. Transfer learning from the pretrained face recognition model proves to be effective for learning deep models for multi-channel PAD task while avoiding overfitting by adapting only a minimal set of DSU's. Overall, the proposed MC-CNN framework uses the information from multiple channels effectively boosting the performance of the system.

The performance breakdown per PAI for BPCER threshold of 1\% is shown in Table \ref{tab:perfbreakdown}. From Table \ref{tab:perfbreakdown} it can be seen that the system achieves perfect accuracy in classifying attacks except for ``glasses''. A discussion about the performance degradation in the ``glasses'' attack is presented in Subsection \ref{subsec:unseen}.


\begin{table}[t]
\centering
\caption{Performance breakdown across different PAIs. Accuracy for each type of attack at at BPCER 1\% is reported.}
\label{tab:perfbreakdown}
\begin{tabular}{@{}cl@{}}
\toprule
ATTACK\_TYPE & MC-CNN @ BPCER : 1\% \\ \midrule
glasses      & 90.82                       \\
fake head     & 100.0                      \\
print        & 100.0                      \\
replay       & 100.0                      \\
rigid mask    & 100.0                    \\
flexible mask & 100.0                      \\
paper mask    & 100.0                       \\
 \bottomrule
\end{tabular}
\end{table}

\begin{table*}[t]
\centering
\caption{Performance of the proposed system as compared to the best baseline method on the $dev$ and $test$ set of the \textbf{grandtest} protocol of WMCA dataset.}
\label{tab:perf_mccn}
\begin{tabular}{@{}ccc|ccc@{}}
\toprule
\multirow{2}{*}{Method} & \multicolumn{2}{c|}{dev (\%)}     & \multicolumn{3}{c}{test (\%)}                                  \\ \cmidrule(l){2-6}
                        & \multicolumn{1}{c|}{APCER} & ACER & \multicolumn{1}{c|}{APCER} & \multicolumn{1}{c|}{BPCER} & ACER  \\ \midrule
\textbf{MC-CNN}  &0.68  &0.84  & 0.6  &   0  &\textbf{0.3}\\
\multicolumn{1}{c}{\begin{tabular}[c]{@{}c@{}}RDWT+Haralick\\  Score fusion\end{tabular}} &6.2  &3.6  &6.39  &0.49  &3.44 \\ 
\multicolumn{1}{c}{\begin{tabular}[c]{@{}c@{}}IQM+LBP\\  Score fusion\end{tabular}}
 &10.52  &5.76  &13.92  &1.17  &7.54 \\ \bottomrule
\end{tabular}
\end{table*}

\subsection{Generalization to unseen attacks}
\label{subsec:unseen}

In this section, we evaluate the performance of the system under unseen attacks. Experiments are done with
the different sub-protocols, which exclude one attack systematically in training. The algorithms are trained with samples from each protocol and evaluated on the $test$ set which contains only the \textit{bonafide} and the attack which was left out in training.
The performance of MC-CNN, as well as the baseline system, are tabulated in Table \ref{tab:unseen_baseline_mccnn}.

\begin{table*}[t]
\centering
\caption{Performance of the baseline and the MC-CNN system with \textbf{unseen attack} protocols. The values reported are obtained with a threshold computed for BPCER 1\% in $dev$ set.}
\label{tab:unseen_baseline_mccnn}
\begin{tabular}{lccc|ccc|lll}
\hline
\multicolumn{1}{c}{\multirow{3}{*}{Protocol}} & \multicolumn{3}{c|}{\begin{tabular}[c]{@{}c@{}}RDWT+Haralick \\ Score fusion\end{tabular}} & \multicolumn{3}{c|}{\begin{tabular}[c]{@{}c@{}}IQM+LBP\\ Score fusion\end{tabular}} & \multicolumn{3}{c}{MC-CNN}     \\ \cline{2-10} 
\multicolumn{1}{c}{}                          & \multicolumn{3}{c|}{test (\%)}                                                             & \multicolumn{3}{c|}{test (\%)}                                                       & \multicolumn{3}{c}{test (\%)} \\ \cline{2-10} 
\multicolumn{1}{c}{}                          & \multicolumn{1}{l|}{APCER}    & \multicolumn{1}{l|}{BPCER}   & \multicolumn{1}{l|}{ACER}   & \multicolumn{1}{l|}{APCER}  & \multicolumn{1}{l|}{BPCER} & \multicolumn{1}{l|}{ACER} & \multicolumn{1}{l|}{APCER}    & \multicolumn{1}{l|}{BPCER}   & \multicolumn{1}{l}{ACER}     \\ \hline
LOO\_fakehead                                 & 4.82                          & 1.5                          & 3.16                        & 4.12                        & 0.64                       & 2.38                      & 0         & 0       & 0       \\
LOO\_flexiblemask                             & 28.06                         & 0.03                         & 14.05                       & 56.36                       & 0.8                        & 28.58                     & 5.04      & 0       & 2.52    \\
LOO\_glasses                                  & 97.09                         & 0.61                         & 48.85                       & 100                         & 1.72                       & 50.86                     & 84.27     & 0       & 42.14   \\
LOO\_papermask                                & 4.01                          & 0.49                         & 2.25                        & 31.51                       & 1.17                       & 16.34                     & 0         & 0.7     & 0.35    \\
LOO\_prints                                   & 0                             & 0                            & 0                           & 3.52                        & 1.08                       & 2.3                       & 0         & 0       & 0       \\
LOO\_replay                                   & 10.03                         & 1.51                         & 5.77                        & 0.15                        & 1.53                       & 0.84                      & 0         & 0.24    & 0.12    \\
LOO\_rigidmask                                & 15.3                          & 0                            & 7.65                        & 27.47                       & 1.08                       & 14.27                     & 0.63      & 0.87    & 0.75    \\ \hline
\end{tabular}
\end{table*}

\begin{figure}[t]
\centering
\includegraphics[width=0.85\linewidth,page=2]{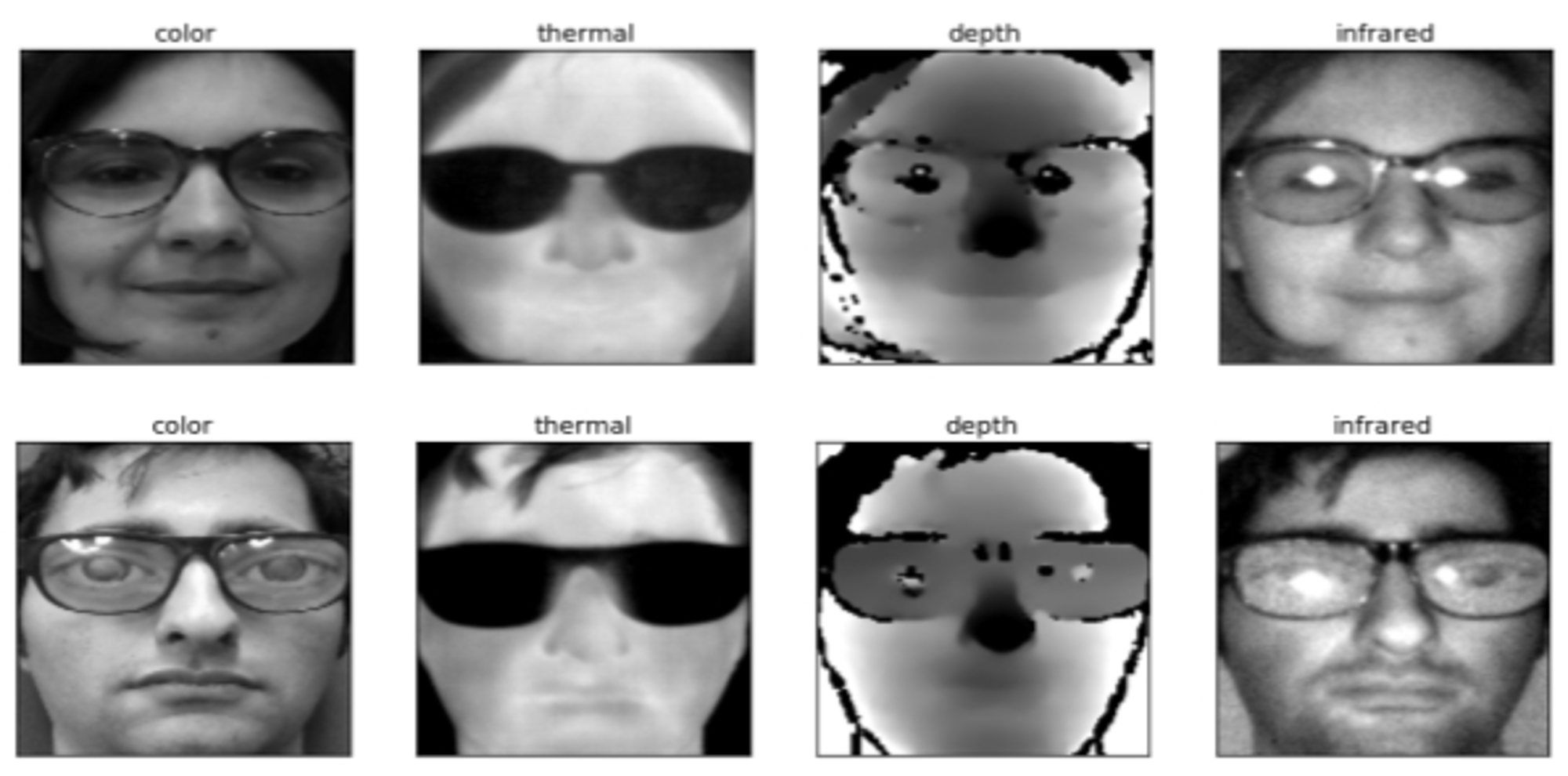}
\caption{Preprocessed data from four channels for \textit{bonafide} with glasses (first row) and funny eyes glasses attack (second row).}
\label{fig:bonafide_funnyeyes}
\end{figure}

From this table, it can be seen that the MC-CNN algorithm performs well in most of the unseen attacks. The baseline methods also achieve reasonable performance in this protocol. However, MC-CNN achieves much better performance as compared to the fusion baselines, indicating the effectiveness of the approach. The performance in the case of ``glasses'' is very poor for both the baseline and the MC-CNN approach. From Figure \ref{fig:bonafide_funnyeyes}, it can be seen that the appearance of the glass attacks are very similar to \textit{bonafide} wearing medical glasses in most of the channels. Since the ``glasses'' attacks were not present in the training set, they get classified as bonafide and reduce the performance of the system.
 
The issue mentioned above is especially crucial for partial attacks in face regions with more variability. For example, partial attacks in eye regions would be harder to detect as there is a lot of variabilities introduced by \textit{bonafide} samples wearing prescription glasses. Similarly, attacks in lower chin could be harder to detect due to variability introduced by \textit{bonafide} samples with facial hair and so on.

\subsection{Analysis of MC-CNN framework}
Here we study the performance of the proposed MC-CNN framework with two different sets of experiments---one with re-training different sets of layers and another with different combinations of channels.

\subsubsection{Experiments with adapting different layers}

Here we try to estimate how adapting a different number of low-level layers affect the performance. The features for the grayscale channel were frozen, and different experiments were done by adapting different groups of low-level layers of the MC-CNN from first layer onwards. In all the experiments, the weights were initialized from the ones trained in the grayscale channel.

Different groups of parameters were re-trained and are notated as follows. The grouping follows the implementation of LightCNN from the authors in the opensource implementation \cite{LightCNN}. The name of the layers is the same as given in 29 layer network described in \cite{wu2018light}.
The notations used for the combination of layers are listed below.
\begin{itemize}
  \item \textbf{FFC} : Only two final fully connected (FFC) layers are adapted.
  \item \textbf{C1-FFC} (1+FFC) : First convolutional layer including MFM, and FFC are adapted.
  \item \textbf{C1-B1-FFC} (1-2+FFC) : Adapting \textit{ResNet} blocks along with the layers adapted in the previous set.
  \item \textbf{C1-B1-G1-FFC} (1-3+FFC) : Adapts \textit{group1} along with the layers adapted in the previous set.
  \item \textbf{1-N+FFC} : Adapts layers from 1 to N along with FFC. 
  \item \textbf{ALL} (1-10 +FFC) : All layers are adapted.
\end{itemize}
Here $FFC$ denotes the two final fully connected layers, and the rest of the names are for different blocks
corresponding to the opensource implementation of LightCNN from \cite{wu2018light}.

The results obtained with re-training different layers are shown in Table \ref{tab:performance_cnn_layers}. 
It can be seen that the performance improves greatly when we adapt the lower layers; however, as we adapt more layers, the performance starts to degrade. The performance becomes worse when all layers are adapted. This can be attributed to over-fitting as the number of parameters to learn is very large. The number of layers to be adapted is selected empirically. The criteria used in this selection is good performance while adapting a minimal number of parameters. For instance, the ACER obtained is 0.3\% for \textit{1-2+FFC} and \textit{1-4+FFC}, the optimal number of layers to be adapted is selected as ``2'' (C1-B1-FFC) since it achieved the best performance adapting a minimal set of parameters. This combination selected as the best system and is used in all the other experiments.

\begin{table*}[t]
\centering
\caption{Performance of the MC-CNN when different combinations of layers were adapted.}
\label{tab:performance_cnn_layers}
\begin{tabular}{@{}ccc|ccc@{}}
\toprule
\multirow{2}{*}{Method} & \multicolumn{2}{c|}{dev (\%)}     & \multicolumn{3}{c}{test (\%)}                                  \\ \cmidrule(l){2-6}
                        & \multicolumn{1}{c|}{APCER} & ACER & \multicolumn{1}{c|}{APCER} & \multicolumn{1}{c|}{BPCER} & ACER  \\ \midrule
FFC (FFC)               &1.51   &1.26  & 2.88  &   0  &1.44 \\
C1-FFC (1+FFC)          &1.77   &1.38  & 2.44  &   0  &1.22 \\
\textbf{C1-B1-FFC(1-2+FFC)}        &0.68   &0.84  &  0.6  &   0  &\textbf{0.3} \\
C1-B1-G1-FFC(1-3+FFC)     &1.1    &1.05  & 1.11  &  0.05  &0.58 \\
C1-B1-G1-B2-FFC(1-4+FFC)  &0.23   &0.61  & 0.58  &  0.02  &0.3 \\
(1-5+FFC)                 &1.14   &0.57  & 0.99  &  0.56  &0.77 \\
(1-6+FFC)                 &100    &50    & 100   &  0   &50 \\
(1-7+FFC)                 &97.56  &48.78 &  96.88&  0   &48.44 \\
(1-8+FFC)                 &99.99  &49.99 & 100 &  0   &50 \\
(1-9+FFC)                 &100    &50    &100    &  0   &50 \\
ALL(1-10+FFC)             &100    &50    &100    &  0   &50 \\ \bottomrule
\end{tabular}
\end{table*}

\subsubsection{Experiments with different combinations of channels}

Here the objective is to evaluate the performance of the algorithm with different combinations of channels.
This analysis could be useful in selecting promising channels which are useful for the PAD task.
Additionally, the performance of individual channels is also tested in this set of experiments to identify the contribution from individual channels. 
It is to be noted that color, depth, and infrared channels are available from the Intel RealSense SR300 device, and the thermal channel is obtained from the Seek Thermal Compact PRO camera. It could be useful to find performance when data from only one sensor is available. We have done experiments
with six different combinations for this task. The combinations used are listed below.
\begin{enumerate}
    \item \textbf{G+D+I+T} : All channels, i.e., Grayscale, Depth, Infrared, and Thermal are used.
    \item \textbf{G+D+I} :  Grayscale, Depth and  Infrared channels are used (All channels from Intel RealSense).
    \item \textbf{G} : Only Grayscale channel is used.
    \item \textbf{D} : Only Depth channel is used.
    \item \textbf{I} : Only Infrared channel is used.
    \item \textbf{T} : Only Thermal channel is used.

\end{enumerate}

The architecture of the network remains similar to the one shown in Fig. \ref{fig:mcnn_general_block}, where only the layers corresponding to the selected channels are present.  Experiments were done using the different combinations of channels with the proposed framework. While training the model, the embeddings from the channels used in a specific experiment are used in the final fully connected layers. The training and testing are performed similarly as compared to experiments conducted in the grandtest protocol.

The results with different combinations of channels are compiled in Table \ref{tab:ablation_channels}. It can be seen that the system with all four channels performs the best with respect to ACER (0.3\%). The combination ``CDI'' achieves an ACER of 1.04\% which is also interesting as all the three channels used is coming from the same device (Intel RealSense). This analysis can be helpful in cases where all the channels are not available for deployment. The performance of the system with missing or with a subset of channels can be computed apriori, and models trained on the available channels can be deployed quickly. Among the individual channels, thermal channel achieves the best performance with an ACER of 1.85\%. However, it is to be noted that the analysis with individual channels is an ablation study of the framework, and the network is not optimized for individual channels. While doing the experiments with individual channels, the architecture is not MC-CNN anymore. The performance boost in the proposed framework is achieved with the use of multiple channels.

\begin{table}[t]
\centering
\caption{Performance of the MC-CNN with various combinations of channels.}
\label{tab:ablation_channels}
\begin{tabular}{@{}ccc|ccc@{}}
\toprule
\multirow{2}{*}{System} & \multicolumn{2}{c|}{dev (\%)}     & \multicolumn{3}{c}{test (\%)}                                  \\ \cmidrule(l){2-6}
                        & \multicolumn{1}{c|}{APCER} & ACER & \multicolumn{1}{c|}{APCER} & \multicolumn{1}{c|}{BPCER} & ACER  \\ \midrule
\textbf{G+D+I+T} &0.68  &0.84  & 0.6  &   0  &\textbf{0.3} \\
G+D+I  &0.78  &0.89  &2.07  &   0  &1.04\\
G  &41.14 & 21.07&65.65 &  0   &32.82\\
D  &10.3  &5.65  &11.77 & 0.31 & 6.04\\
I  & 3.5  &2.25  &5.03  &   0  &2.51\\
T  &4.19  &2.59  &3.14  &0.56  &1.85\\
\bottomrule
\end{tabular}
\end{table}

\subsection{Discussions}

\subsubsection{Performance}

From the experiments in the previous subsections, it can be seen that the performance of the proposed algorithm surpasses the selected feature-based baselines. Transfer learning from face recognition network proves to be effective in training deep multi-channel CNN's with a limited amount of training data.

From the experiments with different channels, it can be seen that the performance of the system with all four channels was the best. We have also tested the same system in a cross-database setting. The data used in this testing was part of the Government Controlled Test (GCT) in the IARPA ODIN \cite{ODIN} project. In the GCT data, it was observed that the system which uses all four channels was performing the best. The addition of complementary information makes the classifier more accurate. The combination of channels makes the framework more robust in general.

The experiments in the unseen attack scenario show some interesting results. Even though the framework is trained as a binary classifier, it is able to generalize well for most of the attacks when the properties of the unseen attacks can be learned from other types of presentations. This can be explained as follows, the 2D PAIs prints and replays can be characterized from depth channel alone. Having one of them in the training set is enough for the correct classification of the other class. The same idea can be extended to other attacks which need information from multiple channels for PAD. For example, if we have silicone masks in the training set; then classifying mannequins as an attack is rather easy.
A PAI is relatively easy to detect when it is distinctive from \textit{bonafide} in at least one of the channels. PAD becomes harder as the representations across channels become similar to that of \textit{bonafide}. This makes the detection of partial attacks such as glasses which occlude a small portion of the face more complex. From the above discussion, it can be seen that, if we have a wide variety of sophisticated attacks in the training set, then the accuracy in detecting simpler unseen attacks seems to be better. This observation is interesting as this could help to tackle the unseen attack scenario, i.e., if we train the system using sufficient varieties of complex PAIs, then the resulting model can perform reasonably well on simpler ``unseen'' attacks. Further, the representation obtained from the penultimate layer of MC-CNN can be used to train one class classifiers/anomaly detectors which could be used to detect unseen attacks.

\subsubsection{Limitations}

One of the main limitations of the proposed framework is the requirement of spatially and temporally aligned channels. Spatial alignment can be achieved by careful calibration of the cameras. Achieving temporal alignment requires the sensors to be triggered in a supervised fashion. However, the proposed framework can handle small temporal misalignments and does not have very stringent requirement on absolute synchronization between channels, as long as there is no significant movement between the frames from different channels. Data from different channels recorded in multiple sessions, as in \cite{agarwal2017face} cannot be used in the proposed framework. In deployment scenarios, the time spent for data capture should be small from the usability point of view; synchronized capture between multiple sensors is suitable for this scenario since it reduces the overall time for data capture. Further, if the multiple channels are not synchronized, face detection in the additional channels is not trivial. Having spatial and temporal alignment obviates the requirement of face detection for all channels since the face location can be shared among different channels. Data capture can be done synchronously as long as the illumination requirements for one sensor is not interfering another sensor and there are no cross sensor interferences. More stringent timing control will be required if there are cross sensor incompatibilities.

From Table \ref{tab:ablation_channels}, it is clear that having multiple channels improves performance significantly. However, it may not be feasible to deploy all the sensors in deployment scenarios. In the absence of certain channels, the proposed framework can be retrained to work with available channels (but with reduced performance). Further, it is possible to extend the proposed framework to work with a different set of additional channels by adding more channels to the proposed framework.

\section{Conclusions}

As the quality of PAIs gets better and better, identifying presentation attacks using visible spectra alone is becoming harder. Secure use of biometrics requires more reliable ways to detect spoofing attempts. Presentation attack detection is especially challenging while presented with realistic 3D attacks and partial attacks. Using multiple channels of information for PAD makes the systems much harder to spoof. In this work, a Multi-channel CNN  framework is proposed, which achieves superior performance as compared to baseline methods.
We also introduce a new multi-channel dataset containing various 2D and 3D attacks tackling identity concealment and impersonation. The proposed database includes a variety of attacks including 2D prints, video and photo replays, mannequin heads, paper, silicone, and rigid masks among others.  From the experiments, it becomes clear that the performance of algorithms is poor when only the color channel is used. Addition of multiple channels improves the results greatly. Furthermore, the unseen attack protocols and evaluations indicate the performance of the system in the real-world scenarios, where the system encounters attacks which were not present in the training set.

\section*{Acknowledgment}

Part of this research is based upon work supported by the Office of the
Director of National Intelligence (ODNI), Intelligence Advanced Research
Projects Activity (IARPA), via IARPA R\&D Contract No. 2017-17020200005.
The views and conclusions contained herein are those of the authors and
should not be interpreted as necessarily representing the official
policies or endorsements, either expressed or implied, of the ODNI,
IARPA, or the U.S. Government. The U.S. Government is authorized to
reproduce and distribute reprints for Governmental purposes
notwithstanding any copyright annotation thereon.

\ifCLASSOPTIONcaptionsoff
  \newpage
\fi

\bibliographystyle{IEEEtran}
\bibliography{refs_mcnn}

\end{document}